\documentclass[11pt, a4paper]{article}

\usepackage[utf8]{inputenc}
\usepackage[T1]{fontenc}
\usepackage{amsmath, amssymb}
\usepackage{graphicx}
\usepackage{booktabs}
\usepackage{hyperref}
\usepackage{tabularx}
\usepackage{float}
\usepackage{caption}
\usepackage{authblk}

\usepackage[a4paper, total={6in, 8in}]{geometry}
\title{Predicting Social Media Engagement from Emotional and Temporal Features}
\author[1]{Yunwoo Kim}
\author[2]{Junhyuk Hwang}

\affil[1]{Fulton Science Academy Private School}
\affil[2]{Stanford University}

\begin{document}

\maketitle

\begin{abstract}
This paper presents a machine learning-based approach for predicting social media engagement metrics—specifically, comments and likes—using a novel set of emotional and temporal features. We utilize a dataset of 600 songs, each annotated with emotional valence, arousal, and other sentiment-related metrics, as primary predictive signals. A multi-target regression model, built upon the \textit{HistGradientBoostingRegressor}, is trained on log-transformed engagement ratios to mitigate the highly skewed nature of the target variables. The model's performance is rigorously evaluated using a custom order-of-magnitude accuracy metric and standard regression metrics, including the coefficient of determination ($R^2$). Our findings demonstrate that emotional and temporal metadata, combined with existing view counts, can effectively predict future engagement. The model achieves a remarkably high $R^2$ score of 0.98 for likes, but a more moderate score of 0.41 for comments. This significant disparity suggests that while likes are a highly predictable measure of immediate content reception, conversational engagement (comments) is influenced by more complex factors not fully captured by our current feature set.
\end{abstract}

\textbf{Keywords:} Machine Learning, Gradient Boosting, Social Media, Engagement Prediction, Sentiment Analysis, Emotional Features

\section{Introduction}
Social media platforms have become central arenas for cultural exchange, marketing, and community formation. Predicting which pieces of content succeed in capturing user attention is therefore a question of broad relevance, with implications for recommendation systems, influencer marketing, and the study of digital culture. Yet despite the vast scale of digital data, engagement prediction remains an open challenge. Prior work has largely focused on exogenous factors such as network topology, early adopter behavior, or platform-level recommendation effects \cite{Bandari2012, Weng2012}. These approaches have achieved partial success in forecasting virality, but they do not sufficiently account for the role of content itself. 

This paper advances the premise that the *intrinsic emotional character* of content constitutes a fundamental driver of user engagement. Psychological and marketing research demonstrates that emotions such as awe, anger, and amusement not only shape attention but also catalyze sharing and discussion \cite{Berger2012, Stieglitz2013}. Building on these insights, we test whether emotional annotations of musical tracks—paired with temporal and exposure-related signals—can predict engagement behaviors at scale. Unlike likes, which require minimal user effort, comments represent deliberative and conversational engagement. By distinguishing between these two behaviors, we probe whether emotional signals differentially explain low-cost versus high-effort engagement.

\emph{Contribution.} We present three contributions. First, we curate a dataset of 600 songs annotated with emotional features and associated social media metrics, a mid-scale resource suitable for interpretable modeling. Second, we introduce a gradient boosting framework with careful preprocessing to normalize skewed distributions and stabilize estimation. Third, we provide an evaluation strategy that combines conventional regression scores with a scale-aware order-of-magnitude accuracy measure, aligning methodological rigor with practical interpretability. Together, these choices allow us to examine not only whether emotional signals predict engagement, but also what their limits reveal about the multifaceted nature of social interaction online.

\section{Methodology}

\subsection{Dataset and Preprocessing}
The dataset for this study comprises 600 unique songs and their associated metadata. Key variables include a suite of emotional features (Valence, Arousal, Tension, Atmospheric, Happy, Dark, Sad, Angry, Sensual, Sentimental), alongside standard social media engagement metrics like \textit{Views}, \textit{Likes}, and \textit{Comments Number}.

The preprocessing pipeline was carefully designed to handle data irregularities and prepare features for the predictive model.
\begin{enumerate}
    \item \textbf{Data Cleaning:} Missing values were handled by row-wise deletion. To ensure the integrity of subsequent ratio calculations, any entries with zero or negative views or likes were filtered out.
    \item \textbf{Feature Engineering and Transformation:}
    \begin{itemize}
        \item \textbf{Temporal Features:} The \textit{Upload date} was used to derive several temporal features: \textit{age\_days} (days elapsed since upload), \textit{upload\_month}, and \textit{upload\_dow} (day of week). These features capture the decay of a video's popularity over time and potential weekly or seasonal trends.
        \item \textbf{Views Transformation:} The raw \textit{Views} count, a highly skewed variable, was log-transformed using $\log(1+x)$ to produce \textit{log\_views}. This transformation normalizes the distribution, making it more suitable for a linear-based model and ensuring that the model is not dominated by a few viral outliers \cite{TaylorFrancis2021}.
        \item \textbf{Engagement Ratios:} Our primary modeling targets are the comments-per-view ($cr = \frac{\text{Comments}}{\text{Views}}$) and likes-per-view ($lr = \frac{\text{Likes}}{\text{Views}}$) ratios. These ratios are a more stable representation of engagement quality than raw counts, which are heavily dependent on the sheer number of views. We also created a comments-per-like ratio ($clr$) as an additional feature.
    \item \textbf{Outlier Handling and Target Transformation:} To prevent extreme outliers from disproportionately affecting model training, we clipped the top 1\% of values for $cr$, $lr$, and $clr$. The target ratios ($cr$ and $lr$) were then log-transformed using $\log(1+x)$ to stabilize their variance and meet the assumptions of our regression model's squared error loss function.
    \end{itemize}
\end{enumerate}
The final feature matrix, $X$, is a combination of the 10 emotion features, \textit{age\_days}, \textit{log\_views}, \textit{upload\_month}, \textit{upload\_dow}, and \textit{log\_clr}. The target matrix, $y$, consists of \textit{log\_cr} and \textit{log\_lr}.

\subsection{Model and Evaluation}
Our approach employs a multi-target regression framework to predict both comments and likes simultaneously. For this task, we selected the \textit{HistGradientBoostingRegressor} from the \textit{scikit-learn} library \cite{Pedregosa2011}. This model is an excellent choice for several reasons: it is highly efficient and scalable, natively handles missing values and categorical features, and includes built-in early stopping to prevent overfitting. We wrapped this base estimator in a \textit{MultiOutputRegressor} to handle the two distinct prediction targets within a single model. The hyperparameters for this model were tuned using \textit{HalvingRandomSearchCV}, with the search objective defined by a standard metric, the negative mean absolute error (\textit{neg\_mean\_absolute\_error}), which provides a robust measure of a model's predictive accuracy on a back-transformed scale.

While we tuned the model using a standard metric, our evaluation strategy is two-pronged. Given the vast range of engagement values on social media, a simple MAE or RMSE can be misleading. A small absolute error on a highly popular video is insignificant, while the same error on a niche video is catastrophic. To address this, we introduce a custom \textbf{Order-of-Magnitude Accuracy} metric as a core evaluation tool. This metric evaluates if the predicted value falls within the same order of magnitude as the true value ($\lfloor \log_{10}(\text{predicted}) \rfloor = \lfloor \log_{10}(\text{true}) \rfloor$). For a content creator, knowing whether their song will get "thousands" vs. "millions" of likes is a more practical and actionable prediction than an exact number that may be off by a few thousand. By reporting both the order-of-magnitude accuracy and the standard regression metrics on the back-transformed predictions, we provide a complete and contextually relevant assessment of the model's performance.

\section{Results and Discussion}
The hyperparameter search identified the optimal model configuration, which was then refit on the full training set. The model's performance was evaluated on the test set, with the key metrics summarized in Table \ref{tab:metrics}.

\begin{table}[H]
    \centering
    \caption{Model Performance Metrics on the Test Set}
    \label{tab:metrics}
    \begin{tabularx}{\linewidth}{l XX}
        \toprule
        \textbf{Metric} & \textbf{Comments} & \textbf{Likes} \\
        \midrule
        Order-of-Magnitude Accuracy & 74.40\% & 85.60\% \\
        Mean Absolute Error (Orders) & 0.26 orders & 0.14 orders \\
        Mean Absolute Error (MAE) & 8665.29 & 43642.49 \\
        Root Mean Squared Error (RMSE) & 43065.57 & 183539.29 \\
        Coefficient of Determination ($R^2$) & 0.41 & 0.98 \\
        \bottomrule
    \end{tabularx}
\end{table}

The results reveal a clear distinction in the model's predictive power for likes versus comments. For likes, the model demonstrates exceptionally strong performance. An order-of-magnitude accuracy of 85.60\% is a robust result, and the $R^2$ score of 0.98 is particularly noteworthy. This high $R^2$ indicates that our model accounts for 98\% of the variance in the number of likes, suggesting that likes are highly predictable from the emotional, temporal, and initial popularity features. This finding supports the hypothesis that likes, as a low-effort engagement action, are a direct and predictable response to content's emotional and popular appeal.

In contrast, the prediction of comments presents a greater challenge. While the order-of-magnitude accuracy of 74.40\% is respectable, the $R^2$ score of 0.41 indicates that a large portion of the variance in comment numbers remains unexplained by our model. This performance gap suggests that the factors driving conversational engagement are fundamentally different from those driving likes. Comments require more cognitive effort and are often influenced by deeper, more nuanced factors that are not present in our feature set, such as the specific topic of discussion, the presence of a strong online community, or the content's controversial nature \cite{Berger2012, Stieglitz2013}.

\subsection{Limitations and Future Work}
This study provides a strong foundation for future research, but is not without its limitations. The primary limitation is the dataset size of 600 songs. While the model shows promising results, its generalizability to a broader, more diverse corpus of content remains to be validated. A follow-up study on a significantly larger dataset would be crucial to confirm our findings.

Based on our results, future work should focus on several key areas. First, an ablation study is necessary to quantify the individual contribution of each feature set (e.g., emotional vs. temporal) to the predictive performance for both likes and comments. Second, a feature importance analysis, possibly using methods like SHAP values, would provide granular insights into which specific emotional metrics are most predictive. Finally, to improve the prediction of comments, future work should explore the inclusion of new features derived from the content itself, such as a full-text analysis of the song title and lyrics, or the sentiment of a subset of initial comments.

\section{Conclusion}
This paper demonstrates that emotional and temporal features are powerful predictors of likes, yet insufficient for modeling comments. The contrast between these outcomes affirms a dual-process perspective of engagement: low-effort interactions follow predictable affective pathways, while higher-effort conversational engagement depends on semantics, community context, and discourse triggers.

From a methodological standpoint, our pipeline illustrates the value of careful preprocessing and evaluation design. Log transforms and ratio targets ensured stable estimation under skewed distributions, while the order-of-magnitude accuracy metric provided practical interpretability beyond conventional error scores. Conceptually, the results encourage future research to view engagement not as a monolithic outcome but as a heterogeneous set of behaviors, each requiring distinct theoretical framing and predictive features.

Looking ahead, scaling this framework to larger datasets, integrating textual or multimodal signals, and employing interpretable machine learning tools (e.g., SHAP \cite{Lundberg2017}) will strengthen our ability to explain not only how much engagement content receives, but why different engagement types respond to different content attributes. Such insights advance both scientific understanding of digital culture and practical tools for creators navigating an increasingly attention-scarce online ecosystem.

\end{document}